# Automated Segmentation for Hyperdense Middle Cerebral Artery Sign of Acute Ischemic Stroke on Non-Contrast CT Images

Jia You[1], Philip L.H. Yu[1], Anderson C.O. Tsang[2], Eva L.H. Tsui[3],
Pauline P.S. Woo[3], Gilberto K.K. Leung[2]

[1] Department of Statistics and Actuarial Science, The University of Hong Kong, Hong Kong
[2] Division of Neurosurgery, Department of Surgery, The University of Hong Kong, Hong Kong
[3] Department of Statistics and Workforce Planning, Hospital Authority, Hong Kong
*plhyu@hku.hk*

**Abstract:**
The hyperdense middle cerebral artery (MCA) dot sign has been reported as an important factor in the diagnosis of acute ischemic stroke due to large vessel occlusion. Interpreting the initial CT brain scan in these patients requires high level of expertise, and has high inter-observer variability. An automated computerized interpretation of the urgent CT brain image, with an emphasis to pick up early signs of ischemic stroke will facilitate early patient diagnosis, triage, and shorten the door-to-revascularization time for these group of patients. In this paper, we present an automated detection method of segmenting the MCA dot sign on non-contrast CT brain image scans based on powerful deep learning technique.

## 1. Introduction:

Acute ischemic stroke (AIS) has becoming a leading cause of morbidity and mortality worldwide and recent advances in endovascular thrombectomy (EVT) for treatment of AIS caused by large vessel occlusion (LVO) have been widely accepted around the world (Powers et al., 2018; Malhotra & Liebeskind, 2015). The hyperdense middle cerebral artery (MCA) dot sign has been reported as an important factor in the diagnosis of acute ischemia, especially in LVO cases (Lim et al., 2018). Fast diagnosis and localization of MCA sign can largely save patients' rescue time, thus lower the probability of severe effect. However, it is fairly challenge to detect MCA sign due to the subtlety of the pathological intensity changes and low signal to noise ratio (Fig. 1). Available data on large vessel occlusion stroke is based on western populations and the respective incidence in Asian countries is largely unknown. So far, this study is the first application of deep learning with specified interest to the hyperdense MCA sign.

We adopted deep learning model as a feature extractor in this study, as well. Recent years saw the availability of large amounts of annotated training sets and the accessibility of affordable parallel computing resources via Graphics Processing Units (or GPUs) have made it feasible to train deep neural networks with huge amount of data and parameters, which have revolutionized the artificial intelligence in achieving the outstanding results on many challenging tasks. The medical imaging community has taken notice of these pivotal developments. The deep learning has been widely applied to medical image analysis, demonstrating the state-of-the-art performances on many medical image analysis tasks, including classification, detection and segmentation.

## 2. Data Preparation:

The Hong Kong Hospital Authority's Clinical Management System (CMS) has well-established records of all patients admitted to the public hospitals for all types of acute ischemic stroke in 2016. The study population was stratified using disproportionate random sampling methods, and the patients' selection criterion is announced in another published paper (Tsang et al., 2019). Total 150 patients were sampled from the ischemic stroke database in CMS. The data includes both CT images and some on-set clinical information, e.g.: side of weakness. The samples were then randomly split into 120 for model training and 30 for validation.

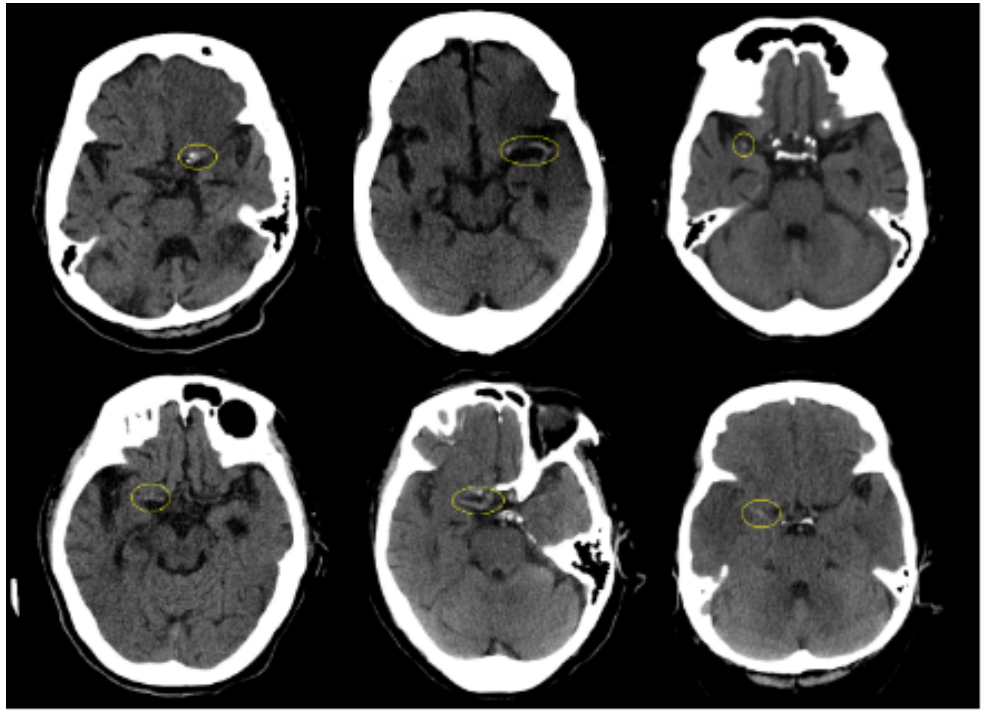

Fig. 1. Sample MCA Dot signs

The MCA ground truth was independently evaluated by two cerebrovascular disease specialists. Any discrepancies were resolved by consensus. The segment labels were manually drawn through software FSL (Smith et al., 2004; Woolrich et al., 2009). Besides imaging data, it also involves the structural data such as patients' side of limb weakness at A & E admission.

The CT images have similar quality, spatial resolution and field- of-view. The in-plane resolution is 0.426*0.426 mm. The slice thickness is 5.0 mm for all cases, and the number of slices is around 26 to 32. Each axial slice has identical resolutions of 512*512.

The existence of hyperdense MCA dot signs can be directly visualized as thromboembolic material within the lumen, which is largely course in a plane perpendicular to the transverse plane of imaging (Fig. 2). Thus the recognition of the MCA dot signs can be localized within a specified area of the scans, and extraction the specified regions of interest will largely help eliminate useless information. We found all MCA dot signs are localized between the 4th and 10th slices after registration to a template. For both training and testing phase, CT scans were pre-processed using the fully automatic pre-processing pipeline through *FSL* and *Nibabel* library under *python 3.5*.

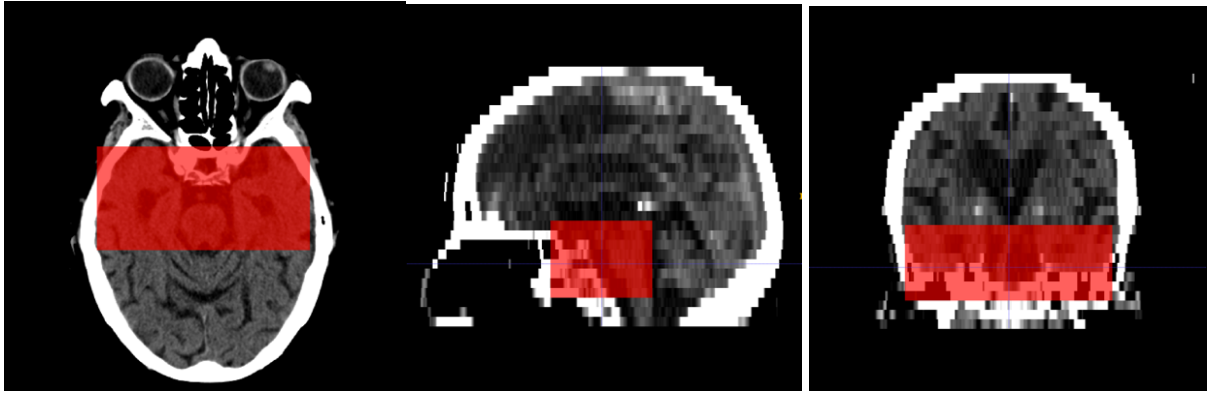

Fig. 2. Regions of Interest for MCA Signs

As shown in pre-processing flow chart (Fig. 3), the first step is brain extraction to strip the skulls. In the second step, all CT scans are rotated and translated through a rigid-body 2D registration procedure in order to make sure all brains within images are horizontally symmetric. All the MCA dot signs have H.U. index between 35 and 60; thus, a threshold of 20 to 80 is utilized in order to eliminate the irrelevant image information and histogram equalization is applied to increase the contrast. To better specify the region where MCA dot sign, we localize a bounding box to subtract the region of interest as Fig. 2. The coloured bounding box has size of 128*128; while two colours indicating left and right hemispheres. The location of MCA within different hemispheres would cause corresponding side of weakness for patients. Given clinical information for different side of limb weakness, we can better localize the infracted hemisphere, coloured in blue and yellow. After extraction of potential infarcted hemisphere, histogram equalization was applied to ROI images to enhance the contrast of MCA dot signs.

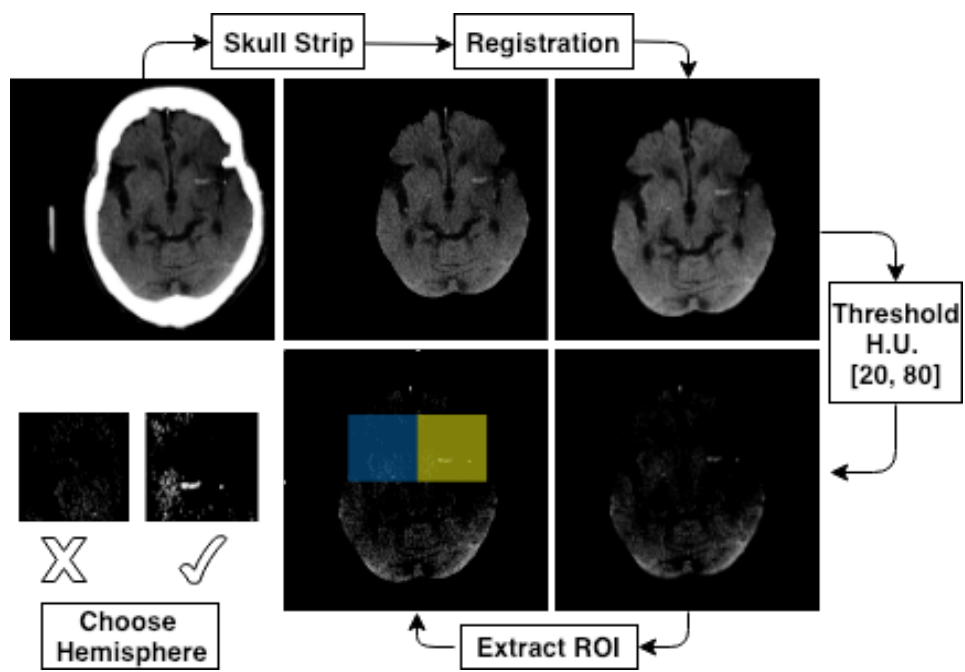

Fig. 3. Pre-processing Flowchart

## 3. Experiments:

The proposed architecture belongs to the category of fully convolutional networks (FCN) (Long & Darrell, 2015) that extends the convolution process across the entire image and predicts the segmentation mask as a whole. This architecture consists of an encoding part and a decoding part, shown as Fig. 4. The encoding part resembles a traditional convolutional neural networks (CNN) (Krizhevsky et al., 2012) that extract a hierarchy of image features from low to high complexity. The decoding part then transforms the features and reconstructs the segmentation label map from coarse to fine resolution. The model contains skip connections, which is pretty similar to the U-net (Ronneberger, et al., 2015), which is one of the most popular architecture for biomedical imaging segmentation tasks. The long-range connections across the encoding part and the decoding part enable high resolution features from the encoding part can be used as extra inputs for the convolutional layers in the decoding part.

Less than half patients, 74 of out 150, in our database has MCA dot sign, and the slice containing ground truth is quite imbalance to empty slices. Due to the limited sample size, we applied data augmentation with randomly zoom in, shift, rotation and horizontal flip of the input images as the final pre-processing step. Moreover, the transfer learning with pre-trained weights could also help during training. Therefore, our encoding structures are exact the same as VGG16 and initial weights are pre-trained on *ImageNet* dataset. Our deep learning model used Adam optimizer with 1e-5 initial learning rate and trained on 200 epochs with Tesla K80 GPU.

To evaluate the performance of the deep learning architecture in segmenting the hyperdense MCA dot signs, the dice similarity coefficients (DSC) is utilized as the evaluation metric for goodness of fit. The DSC is defined as

$$DSC = \frac{2|A \cap B|}{|A| + |B|}$$

where A and B represent the regions of all voxels of ground truth and segmentation respectively.

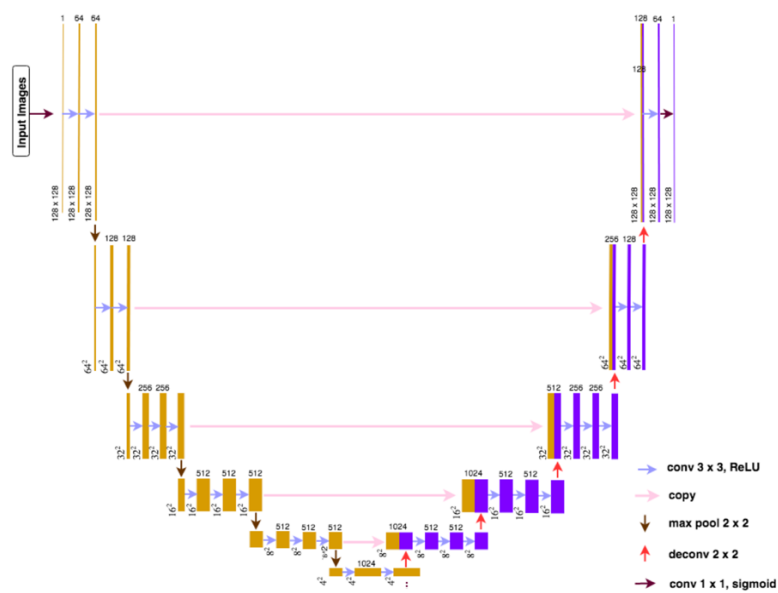

Fig. 4. Deep Learning Architecture

## 4. Result:

Within 150 CT scans, 74 patients were diagnosed to have hyperdense MCA dot sign and the rest were empty. Among the 74 positive MCA sign subjects, 63 were within training and the rest 11 were in the testing. As shown in Table 1, patients without side of weakness do not have MCA dot sign; thus the model involved side of weakness and a filter to select the potential subjects might have MCA.

We subtract cropped bounding boxes within specified regions of interest for each set of CT scans. If patient suffer side of weakness, we extract his or her ROI within corresponding left or right hemisphere; if patient suffer both side of weakness, we use both ROIs in two hemispheres; otherwise, patient without any side of weakness was not considered in model construction. Total Among 120 training subjects, 95 patients (79.16%) were recorded as side of weakness, and 3 patients suffer both sides of weakness. Thus, total 588 slices were fed into model training, and only 84 slices containing MCA sign ground truth.

|  | Side of Weakness | w/o Side of Weakness |  |
|---|---|---|---|
| MCA | 63 | 0 | 63 |
| w/o MCA | 32 | 25 | 57 |
|  | 95 | 25 | 120 |

Table 1. MCA vs Side of Weakness in Training Data

Our model achieves dice similarity coefficient 0.686 on the testing data. The result is satisfied since the MCA sign is extremely small and quite hard to gain relative high DSC.

## 5. Discussion & Conclusion:

MCA dot signs are extreme small in size and quite low signal to noise ratio, the essential step in MCA segmentation task is the localization of specified regions of interest, which would largely ignore those irrelevant information.

The DSC is not high enough mainly from two aspects. The first is due to its size is pretty small that even a subtle miss would cause large effect on the prediction. Total positive ground truth label is less than 0.1% of negative label. Another is the false positive predictions accounts for large inaccuracy that largely due to proximity of bone and the similarity to normal age related vascular calcification. However, our prediction has high sensitivity that able to right predict all MCA dot signs in our testing case.

Further post-processing step to distinguish MCA dot signs and false positive predictions will largely help enhance the model's performance. Adequate data with more positive labelled ground truth will enable model learn more and become more robust, as well.

Overall, we present an automated method for identifying the hyperdense MCA dot sign on Non-contrast CT scans. The study can be further reinforced with additional data input.

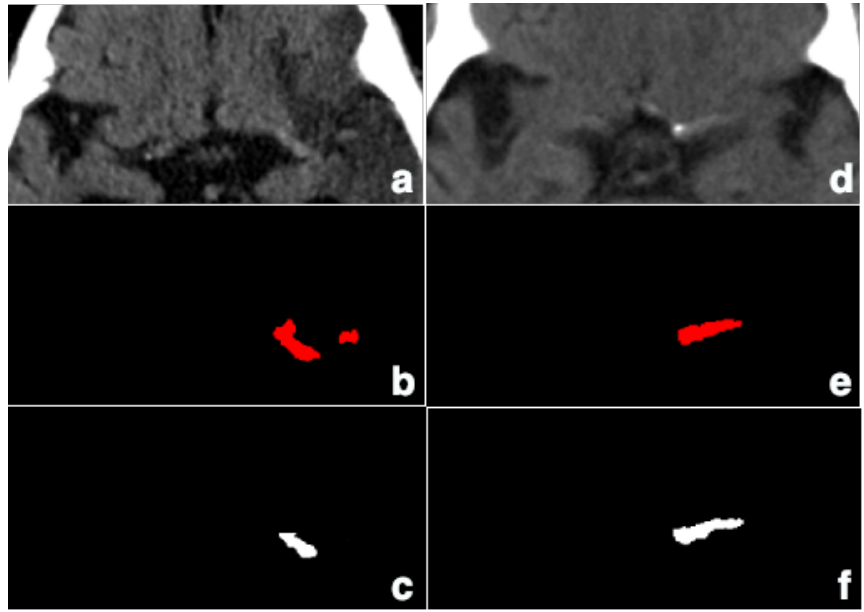

Fig. 4. **a** & **d**: Raw ROIs; **b** & **c**: Ground Truth; **c** & **f**: Predictions

**References:**

Lim, J., Magarik, J. A., & Froehler, M. T. (2018). The CT-Defined Hyperdense Arterial Sign as a Marker for Acute Intracerebral Large Vessel Occlusion. *Journal of Neuroimaging*, *28*(2), 212-216.

Ronneberger, O., Fischer, P., & Brox, T. (2015, October). U-net: Convolutional networks for biomedical image segmentation. In *International Conference on Medical image computing and computer-assisted intervention* (pp. 234-241). Springer, Cham.

Powers, W. J., Derdeyn, C. P., Biller, J., Coffey, C. S., Hoh, B. L., Jauch, E. C., ... & Meschia, J. F. (2015). 2015 American Heart Association/American Stroke Association focused update of the 2013 guidelines for the early management of patients with acute ischemic stroke regarding endovascular treatment: a guideline for healthcare professionals from the American Heart Association/American Stroke Association. *Stroke*, *46*(10), 3020-3035.

Malhotra, K., & Liebeskind, D. S. (2015). Imaging in endovascular stroke trials. *Journal of* neuroimaging, 25(4), 517-527.

Tsang, A. C. O., You, J., Li, L. F., Tsang, F. C. P., Woo, P. P. S., Tsui, E. L. H., … K., G. K. (2019). Burden of large vessel occlusion stroke and the service gap of thrombectomy: A population-based study using a territory-wide public hospital system registry. International Journal of Stroke.

Krizhevsky, A., Sutskever, I., & Hinton, G. E. (2012). Imagenet classification with deep convolutional neural networks. In Advances in neural information processing systems (pp. 1097-1105).

Long, J., Shelhamer, E., & Darrell, T. (2015). Fully convolutional networks for semantic segmentation. In *Proceedings of the IEEE conference on computer vision and pattern recognition* (pp. 3431-3440).

Smith, S. M., Jenkinson, M., Woolrich, M. W., Beckmann, C. F., Behrens, T. E., Johansen-Berg, H., ... & Niazy, R. K. (2004). Advances in functional and structural MR image analysis and implementation as FSL. *Neuroimage*, *23*, S208-S219.

Woolrich, M. W., Jbabdi, S., Patenaude, B., Chappell, M., Makni, S., Behrens, T., ... & Smith, S. M. (2009). Bayesian analysis of neuroimaging data in FSL. *Neuroimage*, *45*(1), S173-S186.